\documentclass{article}
\usepackage{spconf,amsmath,graphicx}
\usepackage{multirow}
\usepackage{booktabs,color,arydshln}
\usepackage{color}
\usepackage{booktabs}
\usepackage{arydshln}
\usepackage{multirow,subfigure}
\usepackage{hyperref,MnSymbol}
\title{Geometry-Aware Video Quality Assessment for Dynamic Digital Human}
\name{Zicheng Zhang$^{1,2}$, Yingjie Zhou$^{1,2}$, Wei Sun$^{1,2}$ ,Xiongkuo Min$^{1,2}$, and Guangtao Zhai$^{1,2,3}$}
\address{$^{1}$Institute of Image Communication and Network Engineering, Shanghai Jiao Tong University, China\\
$^{2}$Peng Cheng Laboratory, China\\
$^{3}$MoE Key Lab of Artificial Intelligence, AI Institute, Shanghai Jiao Tong University, China\\
zzc1998@sjtu.edu.cn}

\begin{document}
%\ninept
%
\maketitle
%

% which have been widely employed in V/AR, game industry, film post-production, etc. Unfortunately, DDHs

\begin{abstract}
Dynamic Digital Humans (DDHs) are 3D digital models that are animated using predefined motions and are inevitably bothered by noise/shift during the generation process and compression distortion during the transmission process, which needs to be perceptually evaluated. Usually, DDHs are displayed as 2D rendered animation videos and it is natural to adapt video quality assessment (VQA) methods to DDH quality assessment (DDH-QA) tasks. However, the VQA methods are highly dependent on viewpoints and less sensitive to geometry-based distortions. Therefore, in this paper, we propose a novel no-reference (NR) geometry-aware video quality assessment method for DDH-QA challenge. Geometry characteristics are described by the statistical parameters estimated from the DDHs' geometry attribute distributions. Spatial and temporal features are acquired from the rendered videos. Finally, all kinds of features are integrated and regressed into quality values. Experimental results show that the proposed method achieves state-of-the-art performance on the DDH-QA database. 
\end{abstract}

\let\thefootnote\relax\footnotetext{This work was supported in part by NSFC (No.62225112, No.61831015), the Fundamental Research Funds for the Central Universities, National Key R\&D Program of China 2021YFE0206700, Shanghai Municipal Science and Technology Major Project (2021SHZDZX0102), and STCSM 22DZ2229005. }

% In practical applications, the DDHs can be degraded by the noise and geometry shift during the scanning generation process. What's more, the DDHs are compressed or simplified during the transmission process under the restriction of bandwidth and storage space.

\section{Introduction}
\label{sec:intro}
With the increasing popularity of digital humans in various applications, such as virtual reality, gaming, and telecommunication, the quality of dynamic digital humans (DDHs) has become a crucial factor in providing a realistic and engaging experience.  To provide useful guidelines for compression algorithms and improve the Quality of Experience (QoE) of viewers, it is necessary to carry out objective quality assessment methods to predict the quality values for DDHs. 
Considering that the DDHs are usually rendered into 2D animation videos for exhibition \cite{zhang2022treating}, it is reasonable to transfer video quality assessment (VQA) methods to DDH quality assessment (DDH-QA) tasks. During the last decade, large amounts of effort have been dedicated to pushing forward the development of VQA. Early full-reference (FR) VQA methods typically use IQA methods, such as PSNR and SSIM \cite{ssim}, to compute the quality difference between reference and distorted frames. Similar to FR-VQA methods, some no-reference (NR) VQA methods compute each frame's quality level using NR image quality assessment (IQA) methods, such as BRISQUE \cite{brisque} and NIQE \cite{niqe}. To incorporate spatial and temporal features, handcrafted-based methods have been proposed, such as VIIDEO \cite{mittal2015completely}, V-BLIINDS \cite{saad2014blind}, TLVQM \cite{korhonen2019two}, and VIDEVAL \cite{tu2021ugc}. Deep neural networks (DNNs) have also been employed, such as VSFA \cite{li2019quality}, RAPIQUE \cite{tu2021rapique}, SimpVQA \cite{sun2022deep}, and FAST-VQA \cite{wu2022fast,wu2022neighbourhood}. 

However, the rendered videos are variant to the viewpoints and the 2D media are not sensitive to the 3D model distortions \cite{zhang2022mm,zhang2023gms}, which indicates simply employing VQA methods is far from enough for DDH-QA task. Therefore, in this paper, we propose a novel NR geometry-aware VQA method to deal with DDH-QA issues. Specifically, geometry attributes including dihedral angle and curvature are computed for the 3D geometry mesh of the digital humans. Then statistical parameters are estimated from the geometry attribute distributions to quantify the geometry distortions such as geometry noise and compression. Afterward, the rendered 2D videos are split into clips for spatial and temporal feature extraction. The first frame of each clip is used for spatial feature extraction with a 2D-CNN backbone while each whole clip is utilized for temporal feature extraction with a fixed pretrained 3D-CNN backbone. The spatial features can help identify the texture distortions like blur and color noise while the temporal features can assist detect motion distortions including motion blur and motion unnaturalness. Later, the geometry, spatial, and temporal features are fused with concatenation and regressed into quality scores with fully-connected (FC) layers.  The experimental results show that the proposed method outperforms all the comparing methods on the DDH-QA \cite{zhang2022ddh} database. 

% To our best knowledge, we are the first to come up with an objective metric specially designed for DDH-QA tasks.

\section{Proposed Method}
\vspace{-0.1cm}
Given a DDH model, we first derive the static digital human geometric mesh $\mathcal{M}$. Afterward, the DDH model is rendered into an animation video sequence $\mathcal{V}$ from a perceptually selected viewpoint, which can cover the major quality information of the DDH model. Both $\mathcal{M}$ and $\mathcal{V}$ are directly provided in the DDH-QA \cite{zhang2022ddh} database. 
\begin{figure}
    \centering
    \includegraphics[width = 0.8\linewidth]{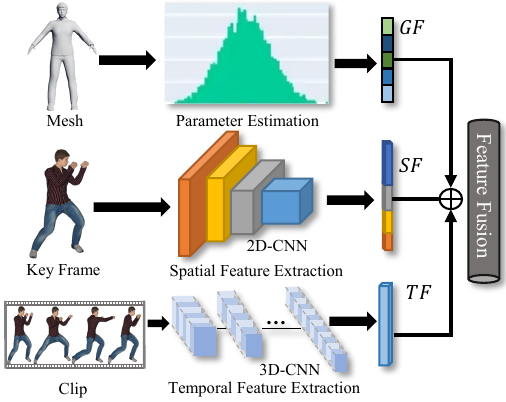}
    \caption{The framework of the proposed method, where the geometry, spatial, and temporal features are extracted from the geometry mesh, key frame, and clip respectively. Then the features are fused with concatenation.}
    \vspace{-0.3cm}
    \label{fig:framework}
\end{figure}
\begin{figure}[!t]
    \centering

    \subfigure[Ref]{\begin{minipage}[t]{0.25\linewidth}
                \centering
                \includegraphics[width=0.6\linewidth,height = 1.9cm]{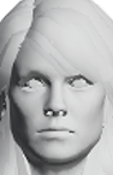}
                %\caption{fig1}
                \end{minipage}}
    \vspace{-0.3cm}
    \subfigure[Ref dihedral angle distribution]{\begin{minipage}[t]{0.7\linewidth}
                \centering
                \includegraphics[width=.9\linewidth,height = 2cm]{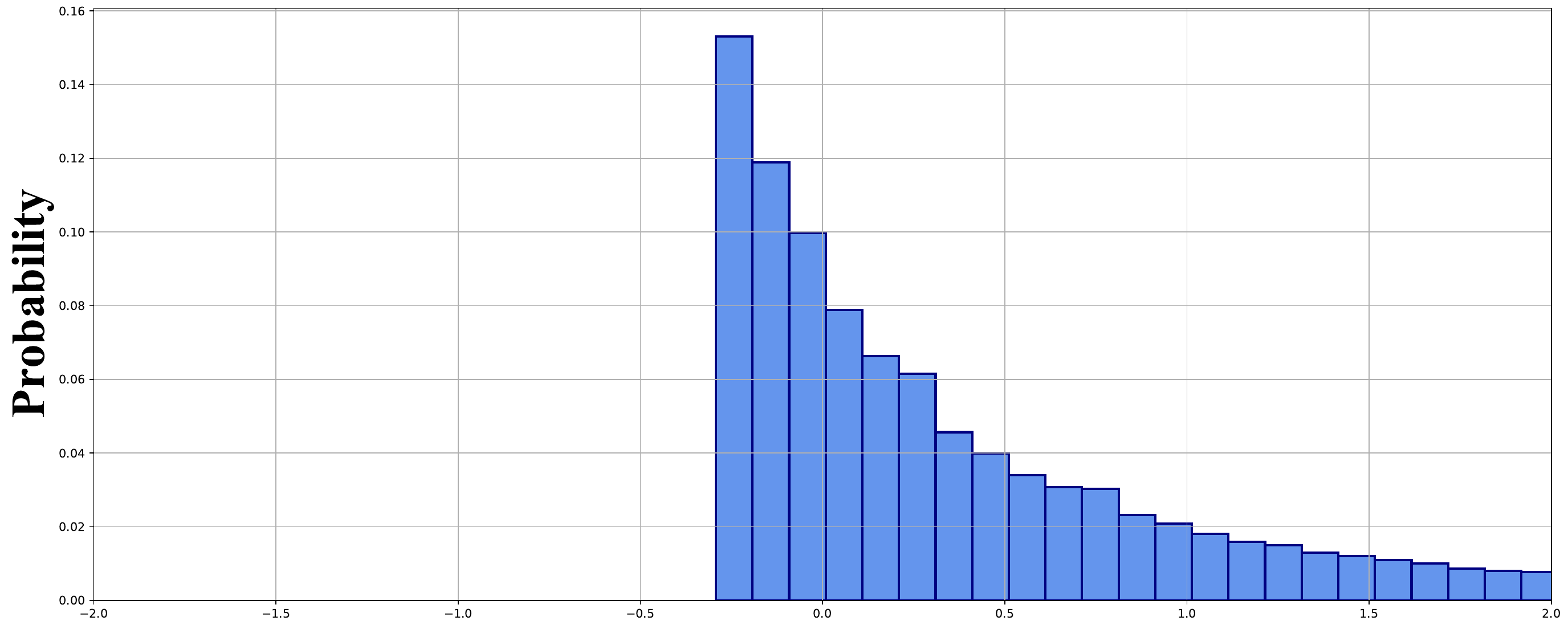}
                %\caption{fig1}
                \end{minipage}}
                
    \subfigure[Compression]{\begin{minipage}[t]{0.25\linewidth}
                \centering
                \includegraphics[width=0.6\linewidth,height = 1.9cm]{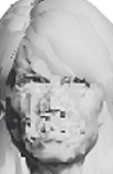}
                %\caption{fig1}
                \end{minipage}}
    \vspace{-0.3cm}
    \subfigure[Compression dihedral angle distribution]{\begin{minipage}[t]{0.7\linewidth}
                \centering
                \includegraphics[width=.9\linewidth,height = 2cm]{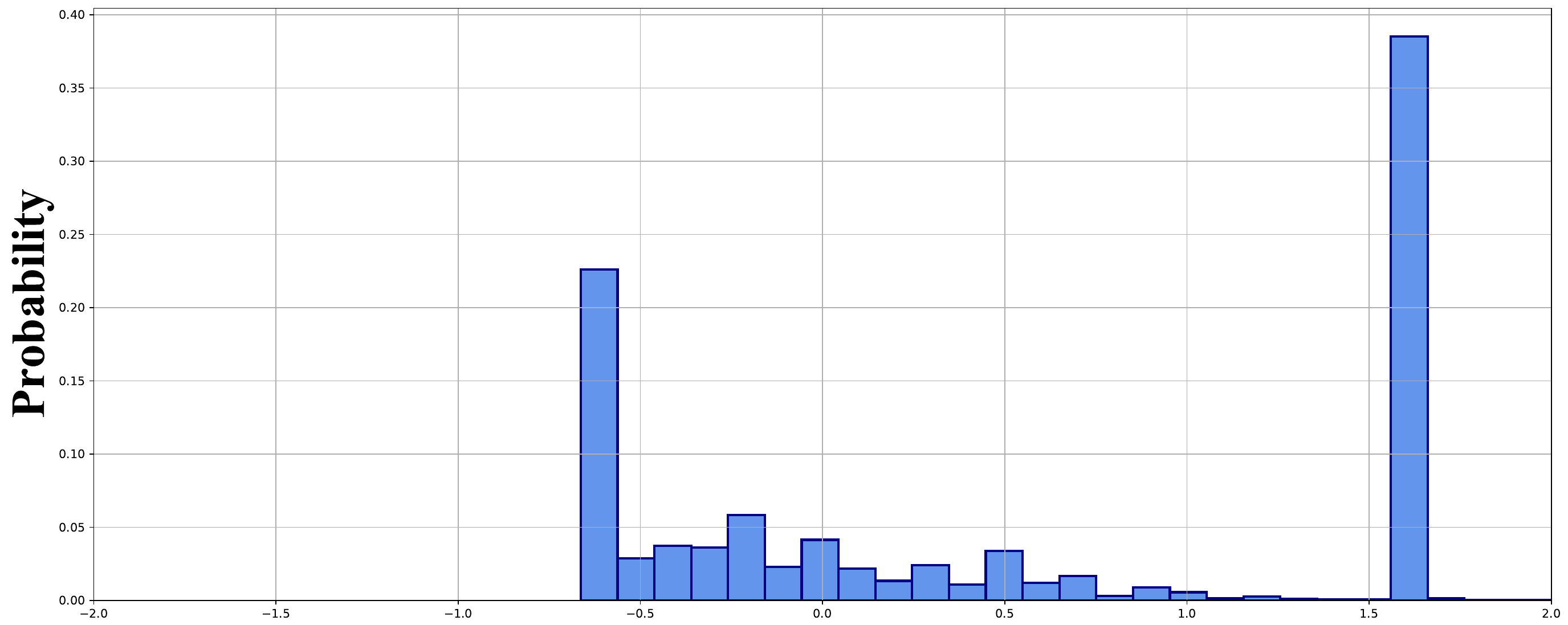}
                %\caption{fig1}
                \end{minipage}}
                
    \subfigure[Ref]{\begin{minipage}[t]{0.25\linewidth}
                \centering
                \includegraphics[width=0.6\linewidth,height = 1.9cm]{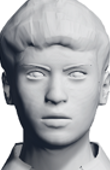}
                %\caption{fig1}
                \end{minipage}}
    \vspace{-0.3cm}
    \subfigure[Ref curvature distribution]{\begin{minipage}[t]{0.7\linewidth}
                \centering
                \includegraphics[width = .9\linewidth,height = 2cm]{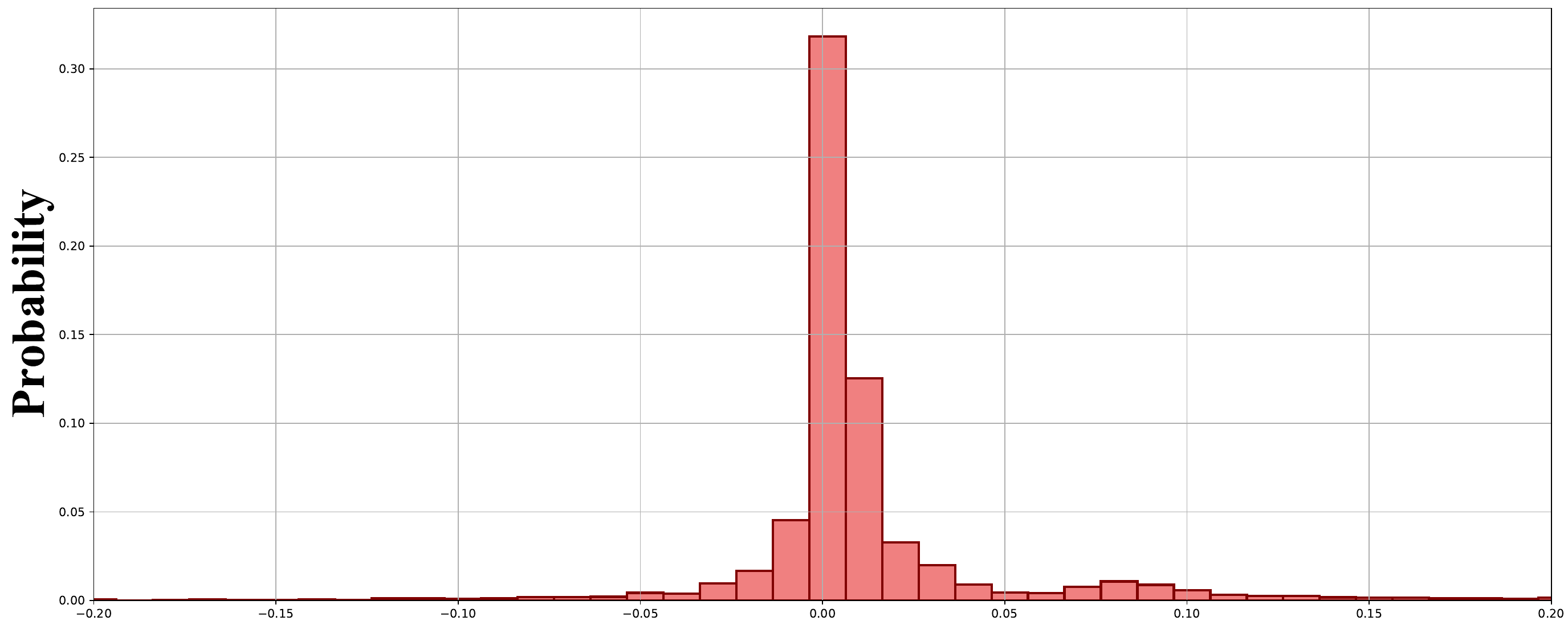}
                %\caption{fig1}
                \end{minipage}}
                
    \subfigure[Noise]{\begin{minipage}[t]{0.25\linewidth}
                \centering
                \includegraphics[width=0.6\linewidth,height = 1.9cm]{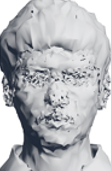}
                \end{minipage}}
    \vspace{-0.3cm}
    \subfigure[Noise curvature distribution]{\begin{minipage}[t]{0.7\linewidth}
                \centering
                \includegraphics[width=.9\linewidth,height = 2cm]{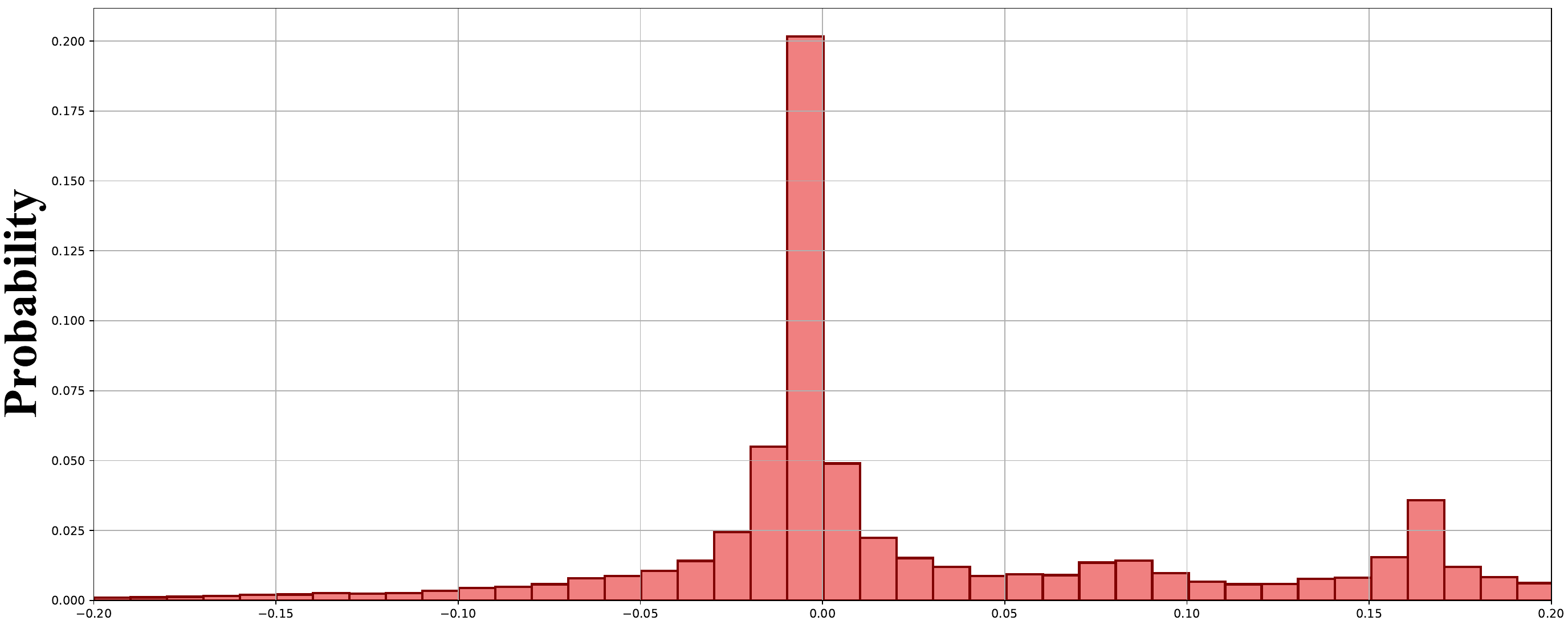}
                %\caption{fig1}}
    \end{minipage}}

    \caption{Reference and distortion examples for the static mesh model along with the corresponding normalized probability distributions for dihedral angle and curvature respectively. The reference distributions can be greatly altered by the presence of distortions. And the estimated statistical parameters are capable of reflecting the perceptual loss from the distribution appearance, as proven in \cite{zhang2022no}.}
    \label{fig:model_distortion}
    \vspace{-0.3cm}
\end{figure}

\vspace{-0.1cm}
\subsection{Geometry Feature Extraction}
\vspace{-0.1cm}
It has been shown in previous works \cite{zhang2022no} that the geometry characteristics are effective for describing the quality-aware local patterns of the 3D models. 
A mesh is typically defined as a collection of vertices, edges, and faces, then we define the geometry mesh for the DDH as:
\begin{equation}
    \mathcal{M} = (\mathbf{Vt}, \mathbf{Eg}, \mathbf{Fc}),
\end{equation}
where $\mathbf{Vt}$, $\mathbf{Eg}$, and $\mathbf{Fc}$ represent the sets of vertices, edges, and faces respectively. 

\vspace{-0.1cm}
\subsubsection{Dihedral Angle}
\vspace{-0.1cm}
The dihedral angle is the angle between the normals of two adjacent faces, which has been regarded as an effective indicator for quantifying the quality of mesh simplification and compression algorithms \cite{angle1}. We can calculate the dihedral angle between two adjacent faces in a mesh by the dot product of corresponding normal vectors:
\begin{equation}
    \cos \theta_{i} = \frac{\mathbf{n}_{i1} \cdot \mathbf{n}_{i2}}{\lVert \mathbf{n}_{i1} \rVert \lVert \mathbf{n}_{i2} \rVert},
\end{equation}
where $\theta_{i}$ denotes the dihedral angle of $i$-th edge $Eg_{i}$, $\mathbf{n}_{i1}$ and $\mathbf{n}_{i2}$ represent the normal vectors of the two adjacent faces whose coedge is $Eg_{i}$. For each edge in the set $\mathbf{Eg}$, its corresponding dihedral angle is computed, which finally generates an array of dihedral angles $\boldsymbol{\theta}$. 

\vspace{-0.1cm}
\subsubsection{Gaussian Curvature}
\vspace{-0.1cm}
Curvature is commonly used for characterizing the features of a surface, such as describing smoothness or roughness, which makes it quite sensitive to structural distortions and thus enables it to assist in describing the visual quality of 3D models \cite{zhang2022no,zhang2021no}. To quantify structural damage for the mesh, we choose Gaussian curvature as the corresponding feature operator:
\begin{equation}
    G_m = \frac{(2\pi - \sum_{n} \theta_{mn})}{A_m},
\end{equation}
where $G_m$ is the Gaussian curvature for the $m$-th vertex $Vt_{m}$, $\theta_{mn}$ is the $n$-th angle between the two adjacent edges at vertex $Vt_{m}$, and $A_i$ indicates the area of the Voronoi cell of vertex $Vt_{m}$. For each vertex in the set $\mathbf{Vt}$, its corresponding Gaussian curvature is computed, which finally generates an array of Gaussian curvature $\mathbf{G}$.

\vspace{-0.1cm}
\subsubsection{Stastical Parameters Estimation}
\vspace{-0.1cm}
The mean, variance, and entropy are employed as the basic statistical parameters. We further choose the generalized Gaussian distribution (GGD) \cite{brisque}, the general asymmetric generalized Gaussian
distribution (AGGD) \cite{brisque}, and the Gamma distribution to estimate quality parameters from the normalized dihedral angle array $\hat{\boldsymbol{\theta}}$ and the normalized Gaussian curvature array $\hat{\mathbf{G}}$.  It has been proven that the appearance of such feature distributions can be altered by various types of distortions, which can be reflected by the estimated statistical parameters \cite{zhang2022no,zhang2021no}. Therefore, these parameters are effective for measuring the visual fidelity of 3D models in the presence of distortion, which can be calculated as:
\begin{equation}
\begin{aligned}
    \mathcal{X} &\sim Basic(\mu,\sigma^{2},E),\\
    \hat{\mathcal{X}} &\sim GGD(\alpha_{1},\beta_{1}), \\
    \hat{\mathcal{X}} &\sim AGGD(\eta, v, \sigma_{l}^{2},\sigma_{r}^{2}), \\
    \hat{\mathcal{X}} &\sim Gamma(\alpha_{2},\beta_{2}),
\end{aligned}
\end{equation}
where $\mathcal{X}$ and $\hat{\mathcal{X}}$ represent the distributions and normalized distributions for dihedral angle and curvature arrays, ($\mu$, $\sigma^{2}$,$E$) parameters in the $Basic(\cdot)$ function stand for (mean, variance, entropy) respectively. 
More specifically, the GGD parameters estimation can be obtained as:
\begin{equation}\small
\begin{aligned} 
&GGD(x;\alpha_{1},\beta_{1}^2)\!=\!\frac{\alpha_{1}}{2 \beta_{1} \Gamma(1 / \alpha_{1})} \exp \! \left(-\left(\frac{|x|}{\beta_{1}}\right)^{\alpha_{1}} \! \right), \label{equ:ggd}\end{aligned}
\end{equation}
where $\beta_{1}=\sigma \sqrt{\frac{\Gamma(1 / \alpha_{1})}{\Gamma(3 / \alpha_{1})}}$, $\Gamma(\alpha_{1})=\int_{0}^{\infty} t^{\alpha-1} e^{-t} dt, \alpha_{1}>0$ is the gamma function, and the two estimated parameters ($\alpha_{1},\beta_{1}^2$) indicate the shape and variance of the distribution. The AGGD parameters estimation can be derived as:
\begin{equation}\small
\begin{aligned}
AGGD(x ; \eta, v, & \sigma_{l}^{2}, \sigma_{r}^{2})= \\
&\left\{\begin{array}{ll}
\!\!\!\frac{v}{\left(\beta_{l}+\beta_{r}\right) \Gamma\left(\frac{1}{v}\right)}\! \exp \! \left(\!-\!\left(\frac{-x}{\beta_{l}}\right)^{v}\right), x<0, \\
\!\!\!\frac{v}{\left(\beta_{l}+\beta_{r}\right) \Gamma\left(\frac{1}{v}\right)}\! \exp \! \left(-\left(\frac{x}{\beta_{r}}\right)^{v}\right), x \geq 0,\\
\end{array}\right.\\
\end{aligned}
\label{equ:aggd1}
\end{equation}
where $\eta$ represents the $\beta_{r}$ and $\beta_{l}$ difference while $\beta_{l} =\sigma_{l} \sqrt{{\Gamma\left(\frac{1}{v}\right)}/{\Gamma\left(\frac{3}{v}\right)}}$ and $\beta_{r} =\sigma_{r} \sqrt{{\Gamma\left(\frac{1}{v}\right)}/{\Gamma\left(\frac{3}{v}\right)}}$, $\sigma_{l}^{2}$ and $\sigma_{r}^{2}$ characterize the spread extent of the distribution on the left and right sides, $v$ determines the shape of the distribution.
The shape-rate Gamma distribution is formulated as:
\begin{equation}
\begin{aligned}
    Gamma(x ; \alpha_{2}, \beta_{2})& =\frac{\beta_{2}^{\alpha_{2}} x^{\alpha_{2}-1} e^{-\beta_{2} x}}{\Gamma(\alpha_{2})} x>0,
\end{aligned}
    \label{equ:gamma}
\end{equation}
where $\alpha_{2}$ and $\beta_{2}$ stands for the shape and rate parameters and $\alpha_{2}, \beta_{2}>0$. In all, a total of 2$\times$(3+2+4+2) = 22 statistical parameters are obtained for describing the geometry perceptual quality for a single DDH and we refer to these features as $GF$.

\vspace{-0.1cm}
\subsection{Video Feature Extraction}
\vspace{-0.1cm}
Given a rendered animation video whose number of frames and frame rate is $n_f$ and $r_f$, we split the video into $\frac{n_f}{r_f}$ clips for feature extraction and each clip lasts for 1s. 

\vspace{-0.1cm}
\subsubsection{Spatial Feature Extraction}
\vspace{-0.1cm}
The spatial features can directly assist the model to identify the existence and extent of common distortions such as blur and noise. Additionally, considering the hierarchical visual perception process, we employ the multi-scale features extracted from a 2D-CNN backbone to incorporate the quality-aware information from low-level to high-level. For the $i$-th clip $C_i$, the first frame is selected as the key frame for spatial feature extraction:
\begin{equation}
\begin{aligned}
     & SF_{i} = \alpha_{1}\oplus \alpha_{2} \oplus \alpha_{3} \oplus \cdots \oplus \alpha_{N_L},  \\
      \alpha_{j} &= {\rm{GAP}}(L_{j}(F_{i})), j \in \{1,2,3,\cdots,N_L\},
\end{aligned} 
\end{equation}
where $SF_{i}$ denotes the extracted spatial features from the key frame of the $i$-th clip, $\oplus$ represents the concatenation operation, $\rm{GAP}(\cdot)$ stands for the global average pooling operation, $L_{j}(F_{i})$ indicates the feature maps obtained from $j$-th layer of the 2D-CNN backbone, $\alpha_{j}$ denotes the corresponding average pooled features, and $N_L$ is the number of the layers for the 2D-CNN. 

\vspace{-0.1cm}
\subsubsection{Temporal Feature Extraction}
\vspace{-0.1cm}
DDHs can be bothered by motion-based distortions such as motion unnaturalness and model clipping. Therefore, to capture the motion-based quality-aware features, we utilize a pretrained 3D-CNN backbone for temporal feature extraction:
\begin{equation}
    TF_{i} = \mathcal{T}(C_i),
\end{equation}
where $TF_{i}$ represents the extracted temporal features from the $i$-th clip $C_i$ and $\mathcal{T}$ indicates the feature extraction operation of the pretrained 3D-CNN backbone.

\vspace{-0.1cm}
\subsection{Feature Fusion \& Quality Regression}
\vspace{-0.1cm}
With the geometry features and video features extracted above, we conduct the clip-level feature fusion by concatenation:
\begin{equation}
    F_i = GF \oplus SF_i \oplus TF_i,
\end{equation}
where $GF$ represents the geometry features for the DDH, $SF_i$ and $TF_i$ indicate the spatial and temporal features extracted from $C_i$, and $F_i$ is the final fused features for $C_i$. Then two-stage fully-connected layers are employed to regress the clip-level features into quality values:
\begin{equation}
    Q_i = FC(F_i),
\end{equation}
where $Q_i$ stands for the predicted quality score for clip $C_i$ and the final quality can be computed via average pooling:
\begin{equation}
    \mathcal{Q} = \frac{1}{N_C} \sum_{i=1}^{N_C} Q_{i},
\end{equation}
where $\mathcal{Q}$ is the final quality score for the DDH and $N_C$ indicates the number of used clips.

\vspace{-0.1cm}
\section{Experiment}
\vspace{-0.1cm}

\begin{table}[t]\small
\renewcommand\arraystretch{0.85}
  \caption{Performance results on the DDH-QA database. Best in {\bf\textcolor{red}{RED}} and second in {\bf\textcolor{blue}{BLUE}}. }
  \vspace{-0.2cm}
  \begin{center}
  \begin{tabular}{c|c|cccc}
    \toprule
    Ref. &  Model & SRCC$\uparrow$ &  PLCC$\uparrow$ & KRCC$\uparrow$ & RMSE$\downarrow$ \\ \hline
    \multirow{2}{*}{FR} 
    &PSNR  &0.4308 &0.5458 &0.3114 &0.9013\\
    &SSIM  &0.5408 &0.6057 &0.3920 &0.8559\\ \hdashline
    \multirow{10}{*}{NR} 
    &BRISQUE &0.3664 &0.4011 &0.2568 &1.0067\\
    &NIQE &0.0923 &0.2489 &0.0748 &1.0418\\

    &VIIDEO &0.1219  &0.1829 &0.0732 &1.0740\\
    &V-BLIINDS &0.4807  &0.4936 &0.3424 &0.9564\\
    &TLVQM &0.2515  &0.2824 &0.1729 &1.0480\\
    &VIDEVAL &0.2218  &0.3470 &0.1622 &1.0246\\
    &VSFA &0.5406  &0.5708 &0.3858 &0.9657\\
    &RAPIQUE &0.1815  &0.2368 &0.1246 &1.0614\\
    &SimpVQA &\bf\textcolor{blue}{0.7444}  & \bf\textcolor{blue}{0.7498} & \bf\textcolor{blue}{0.5452} & \bf\textcolor{blue}{0.7228}\\
    &FAST-VQA &0.5262  &0.5382 &0.3657 &1.0499\\
    &Proposed &\bf\textcolor{red}{0.8004} &\bf\textcolor{red}{0.7956} &\bf\textcolor{red}{0.6028} &\bf\textcolor{red}{0.6343}\\

    \bottomrule
  \end{tabular}
  \end{center}
  \vspace{-0.7cm}
  \label{tab:performance}

\end{table}

\vspace{-0.1cm}
\subsection{Experimental Setup \& Implementation Details}
\vspace{-0.1cm}
The DDH-QA \cite{zhang2022ddh} database is employed for validation, which provides 800 degraded DDHs with both model-based and motion-based distortions.
The DDH videos are with varying durations (2s$\sim$8s) and are divided into 1s clips. We uniformly employ 6 clips from each video with cyclic sampling. Specifically, if a video has less than 6 clips, the existing clips are expanded with cyclic sampling until 6 clips are selected. For videos lasting for more than 6 seconds, the first 6 clips are used. The ResNet50 \cite{he2016deep} is employed as the spatial feature extractor and patches with a resolution of 448$\times$448$\times$3 are cropped as input. The SlowFast R50 \cite{feichtenhofer2019slowfast} is utilized as the temporal feature extractor and the clips are resized to 224$\times$224 for both training and testing. The ResNet50 is initialized with a pre-trained model on the ImageNet database \cite{deng2009imagenet} and fine-tuned during the training phase. While the SlowFast R50 is frozen, with pre-trained model weights on the Kinetics 400 database \cite{kay2017kinetics}. The Adam optimizer \cite{kingma2014adam} is utilized, with an initial learning rate of 4e-6. The default number of epochs and batch size are set as 30 and 4. The mean squared error (MSE) is used as the loss function.

The 5-fold cross-validation strategy is employed. In this strategy, the 10 motion groups are split into 5 folds, with each fold containing 2 groups of motion. Four folds are utilized as training sets, while the remaining fold is used as the testing set. This process is repeated 5 times, ensuring that every fold is used as the testing set. The final experimental results are obtained by recording the average performance. Furthermore, for methods that do not require training, we apply them to the same testing sets and report their average performance.

\vspace{-0.1cm}
\subsection{Benchmark Competitors \& Criteria}
\vspace{-0.1cm}
To validate the animated videos in the DDH-QA database, several video quality assessment (VQA) methods are utilized. The FR methods, such as PSNR and SSIM \cite{ssim}, operate on the frame level of the DDH videos. The NR methods include handcrafted-based methods, such as BRISQUE \cite{brisque}, NIQE \cite{niqe}, VIIDEO \cite{mittal2015completely}, V-BLIINDS \cite{saad2014blind}, TLVQM \cite{korhonen2019two}, and VIDEVAL \cite{tu2021ugc}, as well as DNN-based methods, such as VSFA \cite{li2019quality}, RAPIQUE \cite{tu2021rapique}, SimpleVQA \cite{sun2022deep}, and FAST-VQA \cite{wu2022fast}. 

Four mainstream consistency evaluation criteria are utilized to compare the correlation between the predicted scores and MOSs, which include Spearman Rank Correlation Coefficient (SRCC), Kendall’s Rank Correlation Coefficient (KRCC), Pearson Linear Correlation Coefficient (PLCC), and Root Mean Squared Error (RMSE).

\begin{table}[!t]\small
\centering
\renewcommand\arraystretch{0.85}
\caption{Experimental performance of the ablation study, where GF, SF, and TF indicate the geometry features, spatial features, and temporal features respectively.}
\setlength{\tabcolsep}{2pt}
\begin{tabular}{c|cccc}
\toprule
Feature &  SRCC$\uparrow$     & PLCC$\uparrow$    & KRCC$\uparrow$   &RMSE$\downarrow$ \\\hline
GF+SF &0.7771 &\bf\textcolor{blue}{0.7762} &\bf\textcolor{blue}{0.5896} &0.6888\\
GF+TF &0.4312 &0.4903 &0.2961 &0.9393 \\
SF+TF &\bf\textcolor{blue}{0.7786} &0.7731 &0.5860 &\bf\textcolor{blue}{0.6702}\\ \hline
All &\bf\textcolor{red}{0.8004} &\bf\textcolor{red}{0.7956} &\bf\textcolor{red}{0.6028} &\bf\textcolor{red}{0.6343}  \\  \bottomrule
\end{tabular}

\label{tab:ablation}
\vspace{-0.5cm}
\end{table}

\begin{table}[!t]\small
\centering
\renewcommand\arraystretch{1}
\caption{SRCC Experimental performance corresponding to the used number of clips. Since the videos last for 2s$\sim$8s, we test the proposed method with numbers of clips from 2$\sim$8.}
\setlength{\tabcolsep}{2pt}
\begin{tabular}{c|ccccccc}
\toprule
Num &2 &3 &4 &5 &6 &7 &8\\\hline
SRCC &0.6914 &0.7150 &7501 & 0.7711 & \bf\textcolor{red}{0.8004} & \bf\textcolor{blue}{0.7850} & 0.7745\\  \bottomrule
\end{tabular}

\label{tab:clip}
\vspace{-0.5cm}
\end{table}

\vspace{-0.1cm}
\subsection{Performance Discussion}
\vspace{-0.1cm}
The experimental performance is exhibited in Table \ref{tab:performance}, from which we can see that the proposed method outperforms all the compared methods and surpass the second-place method SimpVQA by about 7.5\% in terms of SRCC, which indicates the effectiveness of the proposed method for evaluating the perceptual quality of DDHs. To further investigate the contributions of different types of features, we conduct the ablation study and the results are shown in Table \ref{tab:ablation}. With closer inspection, we can find that using all three types of features achieves the best performance, which reveals that GF, SF, and TF make contributions to the final results. Moreover, we test the influence of utilizing varying numbers of clips and the results are illustrated in Table \ref{tab:clip}. It can be seen that when using smaller than 6 clips, the performance can be improved by the increasing number of clips. However, using 7 or 8 clips can cause performance drops due to redundancy and over-fitting.

\vspace{-0.1cm}
\section{Conclusion}
\vspace{-0.1cm}
In conclusion, this paper proposes a novel no-reference geometry-aware video quality assessment method for Dynamic Digital Humans. By leveraging statistical parameters estimated from DDHs' geometry attribute distributions and spatio-temporal features acquired from rendered videos, the proposed method achieves state-of-the-art performance on the DDH-QA database. The approach offers an effective solution to the DDH quality assessment (DDH-QA) tasks, which can be useful for improving the visual quality of DDHs and enhancing the user experience in various applications.
\bibliographystyle{IEEEbib}
\bibliography{strings,refs}

\end{document}